\documentclass[runningheads]{llncs}
\usepackage{makeidx}  
\usepackage{amssymb}
\usepackage{graphicx}
\usepackage{epsfig}
\usepackage{amsmath}
\usepackage{array}
\usepackage[T1]{fontenc}
\newcolumntype{M}[1]{>{\centering\arraybackslash}m{#1}}
\newcolumntype{L}[1]{>{\flushleft\arraybackslash}m{#1}}

\makeatletter
\g@addto@macro\normalsize{%
  \setlength\abovedisplayskip{4pt}
  \setlength\belowdisplayskip{4pt}
  \setlength\abovedisplayshortskip{4pt}
  \setlength\belowdisplayshortskip{4pt}
}
\makeatother

\begin{document}
\title{Surgical Skill Assessment on In-Vivo Clinical Data via the Clearness of Operating Field}
\titlerunning{Surgical Skill Assessment}

\author{
Daochang Liu\inst{1} \and
Tingting Jiang\inst{1} \and
Yizhou Wang\inst{1,3,4} \and
Rulin Miao\inst{2} \and
Fei Shan\inst{2} \and
Ziyu Li\inst{2}}


\authorrunning{D. Liu et al.}
%
\institute{
NELVT, Department of Computer Science, Peking University, China \and
Peking University Cancer Hospital \and
Peng Cheng Lab \and 
Deepwise AI Lab \\ 
\email{\{daochang, ttjiang\}@pku.edu.cn} \\
 }

\maketitle              

\begin{abstract}
Surgical skill assessment is important for surgery training and quality control. 
Prior works on this task largely focus on basic surgical tasks such as suturing and knot tying performed in simulation settings. 
In contrast, surgical skill assessment is studied in this paper on a real clinical dataset, which consists of fifty-seven in-vivo laparoscopic surgeries and corresponding skill scores annotated by six surgeons. 
From analyses on this dataset, the clearness of operating field (COF) is identified as a good proxy for overall surgical skills, given its strong correlation with overall skills and high inter-annotator consistency. 
Then an objective and automated framework based on neural network is proposed to predict surgical skills through the proxy of COF. 
The neural network is jointly trained with a supervised regression loss and an unsupervised rank loss. 
In experiments, the proposed method achieves 0.55 Spearman's correlation with the ground truth of overall technical skill, which is even comparable with the human performance of junior surgeons. 

\end{abstract}

\section{Introduction} 
Surgical skill assessment is crucial for the improvement of surgeons' competency during both training and practice~\cite{2017Survey}. 
Traditionally, surgical skills are evaluated by experts with onsite observation, which is prone to subjective biases. 
Although OSATS~\cite{OSATS} can reduce such subjectivity to some extent, it requires intensive efforts from experts for manual grading. 
Thus computer-aided approaches, which provide objective and scalable skill predictions, draw increasing attention from the community. 
This paper works on computer-aided surgical skill assessment.

Prior works on this task~\cite{2018MICCAI,2018IJCARS_Zia_1,2018IJCARS_Wang,2018IJCARS_Zia_2,2018CVPR,2018IJMRCAS,2016MICCAI,2016IJCARS,2015TPAMI,2014ISBI,2013MICCAI} have been mainly set up on simulated datasets such as JIGSAWS~\cite{JIGSAWS}, in which basic suturing/tying tasks are performed on benchtop models. 
Although some benchtop models can have high fidelity of human anatomy and accurate imitation of procedure steps, there is still a large gap between simulated scenarios and real-world clinical ones in terms of varying patient condition, working dynamics, stress level and so on. 
Only few studies~\cite{2019AnnS,2018WACV,2015IJCARS,2010AnnS} have been conducted on clinical data. 
As for surgery type, clinical open surgeries have been studied in~\cite{2019AnnS,2015IJCARS} and laparoscopies in~\cite{2018WACV,2010AnnS}. 
The modality of these clinical datasets can be surgery videos~\cite{2019AnnS,2018WACV} or the data from external sources~\cite{2015IJCARS,2010AnnS}.
The limitations of the two existing clinical video datasets are that~\cite{2019AnnS} only targets at short suturing/tying clips manually segmented from long surgeries and~\cite{2018WACV} annotates only four videos with skill labels. 
Different from previous works, we construct a new clinical laparoscopic video dataset, which consists of \textit{fifty-seven long} procedures.





On the other hand, from the perspective of assessment method, most of the existing works are motion-based. 
In these works, surgical skills are determined with hand/tool/eye motions, obtained from robotic kinematics~\cite{2018MICCAI,2018IJCARS_Wang,2018IJCARS_Zia_2,2018IJMRCAS,2013MICCAI} or external sensors~\cite{2018IJCARS_Zia_1,2016MICCAI,2015IJCARS,2010AnnS} or visual tracking/interest points~\cite{2019AnnS,2018WACV,2018IJCARS_Zia_1,2016IJCARS,2015TPAMI,2014ISBI}. 
However, acquiring motion trajectories is difficult for clinical data because that 1) robotic kinematics are restricted to only robotic surgeries 2) external sensors interrupt normal workflows and are hard to implement in operating rooms 3) visual tracking is not robust enough and often involves manual corrections. 
Unlike these prior works, we utilize the clearness of operating field (COF) rather than motion for skill assessment. 
The COF reflects the amount of bleeding and the visibility of anatomy landmarks.
Prediction of the COF relies on appearance information carried in surgery videos, which is more obtainable and robust than motions in clinical settings. 
Statistical analyses on our clinical dataset identify that the COF is a good skill proxy, for its strong correlation with overall skills and high consistency across annotators.
For a detailed review of previous studies, please refer to this survey~\cite{2017Survey}. 

In this paper, a new clinical dataset is collected, which includes fifty-seven videos of laparoscopic gastrectomy conducted in operating rooms (OR). 
Six surgeons annotate the videos with not only technical OSATS scores but also newly designed procedural scores and the COF score.  
Then we propose an objective video-based method to predict surgical skills via the proxy of COF. 
A neural network taking in color and semantic features is trained with a supervised regression loss and an unsupervised rank loss collectively. 
The proposed method outputs skill scores at frame-level to provide feedback. 
Experiments show that predicting overall skills via the proxy of COF performs better than predicting them directly without proxy. 
The Spearman's correlation between predicted overall technical skill and ground truth ratings is 0.55, which is promising and even comparable with the human performance of junior surgeons. 
In summary, our contributions are three-fold: 1) An in-vivo clinical dataset collected from real operating rooms. 2) The identification of COF as a good skill proxy. 3) A video-based method without extra equipment and intensive human efforts. 


\section{Dataset}

\begin{table}[t]
\begin{center}
\caption{Skill Metrics}
\label{table:table1}
\small
\begin{tabular}{M{0.35cm} M{5.1cm}||M{0.35cm} M{5.7cm}}
\hline
\hline
ID & {\bf General Metrics} & ID & {\bf Procedure-specific Metrics}\\
\hline
\hline
1 & Gentleness & 7 & Dissection in Correct Planes\\
2 & Time and Motion & 8 & Vessel Exposure and Transection\\
3 & Instrument Handling & 9 & Venous Breakpoint Selection\\
4 & Flow of Operation & 10 & Arterial Breakpoint Selection\\
5 & Tissue Exposure & 11 & Infrapyloric Artery Exposure\\
6 & Overall Technical Skill (OTS) & 12 & Care for Adjacent Organs\\\cline{1-2}
14 & Clearness of Operating Field & 13 & Overall Procedural Skill (OPS)\\   
\hline
\hline
\end{tabular}
\end{center}
\end{table}

\begin{figure}[t]
\begin{center}
   \includegraphics[width=1.0\linewidth]{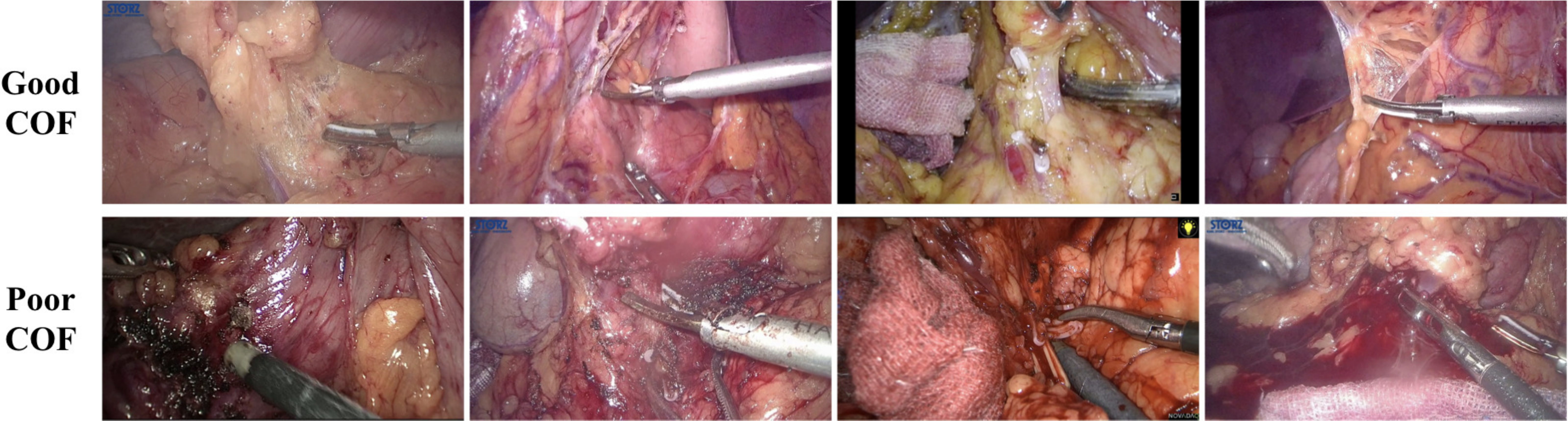}
\end{center}
   \caption{Example frames from cases with good or poor COF scores.}
\label{fig:COF_example}
\end{figure}

{\bf Data.} 
The dataset includes videos of 57 in-vivo laparoscopic gastrectomy surgeries for gastric cancer. 
Videos are captured by the built-in camera (Karl Storz, Olympus or Sony) and are formatted to 960$\times$540 and 25 fps. 
The procedures are performed by one surgeon. 
It is assumed that the performance of a same primary surgeon can vary across cases, due to different case complexity, fatigue condition, team members, operating time and so on. 
To lighten the annotation burden, the infrapyloric area~\cite{GastrectomyBook}, which is one of the four major parts of gastrectomy, is used for skill assessment in this study. 
The duration of infrapyloric procedure ranges from 8 to 57 minutes and the average is 26 minutes. 
Surgical skills are annotated by 6 surgeons on 14 metrics with the Likert scale of 1-5. 
The 6 surgeons include 3 senior surgeons with more than 8 years experience and 3 junior surgeons with less than 4 years experience. 
For each metric, the ground truth is defined as the mean score of the three seniors. 

\noindent
{\bf Skill Metrics.} 
As listed in Table \ref{table:table1}, the surgeries are evaluated on 14 metrics, including technical OSATS metrics, procedural metrics, and the COF metric. 
For OSATS metrics (ID 1-6), a modified version from~\cite{NewEngland} is employed. 
Since OSATS metrics only measure the general surgical technique and are procedure-independent, we also design 7 new procedure-specific metrics (ID 7-13) according to~\cite{GastrectomyBook} to provide fine-grained ratings of surgeons' compliance with gastrectomy instructions. 
In addition, based on the suggestion of surgeons, the COF metric (ID 14) is proposed to represent how much the bleeding and burned tissues impact on the identification of anatomical structures. 
Examples are shown in Fig. \ref{fig:COF_example}. 
In cases with good COF, there is no obvious bleeding and the anatomical planes and landmarks can be recognized clearly, while in poor cases the bleeding is excessive and severely affects the recognition of planes and landmarks.



\begin{figure}[t]
\begin{center}
   \includegraphics[width=1.0\linewidth]{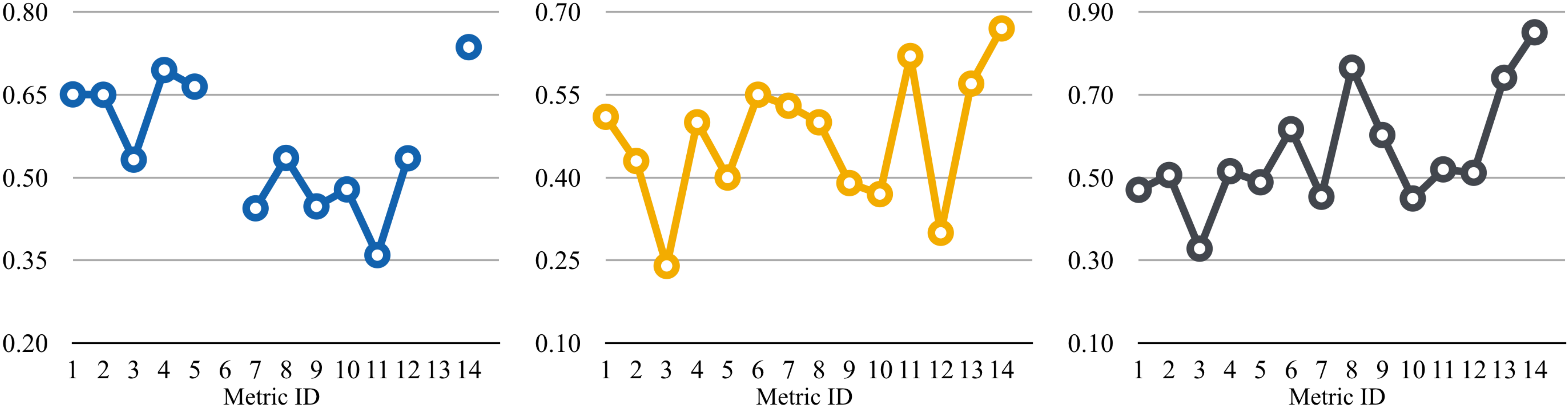}
\end{center}
   \caption{Metric analyses. {\bf Left:} Correlation with overall skills. {\bf Middle:} Inter-senior consistency. {\bf Right:} Senior-junior consistency. COF (ID 14) is a good skill proxy.}
\label{fig:dataset_analysis}
\end{figure}

\noindent
{\bf Analysis.} 
As shown in Fig. \ref{fig:dataset_analysis}, skill metrics are analyzed statistically on all 57 cases in terms of correlation with overall surgical skills and inter-annotator consistency. In the following analyses, ${\mathbf r}_{im}$ denotes the 57 scores from surgeon $i$ on metric $m$. $\mathcal{S}$ and $\mathcal{J}$ denote the sets of seniors and juniors respectively. And $\mathrm{srocc}(\cdot,\cdot)$ is the function of Spearman's rank correlation coefficient (SROCC).

\textit{Correlation with overall skills.} We consider the correlations with both overall technical skill (OTS) and overall procedural skill (OPS). 
Concretely, the correlation with overall skills for a non-overall metric $m$ is defined as:
\begin{equation}
{\textstyle CORR_m = \frac{1}{2\times|\mathcal{S}|} \sum_{i \in \mathcal{S}} \sum_{k \in \{6,13\}} \mathrm{srocc}({\mathbf r}_{im},{\mathbf r}_{ik}) \ \ \ m \notin \{6,13\}}
\end{equation}
where the SROCC between this metric and either OTS or OPS is computed, and the results from three seniors are averaged.
This value is the largest when $m=14$, indicating that COF has the best correlation with overall skills.

\textit{Inter-senior consistency.} 
This value reflects how seniors agree on a metric. 
Specifically, the inter-senior consistency for metric $m$ is defined as the averaged SROCCs between the scores of this metric from each two seniors:
\begin{equation}
{\textstyle ISC_m = \frac{1}{|\mathcal{S}|\times(|\mathcal{S}|-1)} \sum_{i,j \in \mathcal{S}, i \neq j} \mathrm{srocc}({\mathbf r}_{im},{\mathbf r}_{jm}).}
\end{equation}
The higher this value is, the metric is less affected by subjective biases.
Among the 14 metrics, the COF achieves the highest inter-senior consistency. 

\textit{Senior-junior consistency.} To examine whether juniors and seniors have a similar understanding of each metric $m$, 
we define senior-junior consistency as the SROCC between the ground truth and the mean scores of the three juniors:
\begin{equation}
{\textstyle SJC_m = \mathrm{srocc}(\frac{1}{|\mathcal{S}|} \sum_{i \in \mathcal{S}} {\mathbf r}_{im}, \frac{1}{|\mathcal{J}|} \sum_{j \in \mathcal{J}} {\mathbf r}_{jm}).}
\end{equation}
A high value means that this metric can be correctly understood by juniors and can thus provide effective feedback to juniors. 
The COF gives the best value. 

In addition to the above reasons, the assessment of COF only relies on appearance information carried in video data. 
Hand/tool motions and extra devices are not involved. 
Therefore the COF is identified as a good proxy for overall skills.



\section{Method}

In this section, an objective video-based method is devised to regress surgical skills via the proxy of COF. 
The proposed approach takes a video as input and predicts its COF score, which is directly regarded as overall surgical skills.

\noindent
{\bf Preprocessing.} 
The input video is first downsampled to 1 fps to reduce data redundancy. 
Then extra-abdominal views are removed manually. 
Note that this can be efficiently done by non-professionals.

\noindent
{\bf Feature Extraction.} 
Since the COF is a measure for bleeding amount and anatomical recognition, we extract color features to describe the severity of bleeding and semantic features to provide anatomy information. 
For color features, color histograms in RGB space, HSV space, and Red-ratio space (R/G and R/B) are computed in every video frame. 
In our dataset, videos can have inconsistent color distributions due to different recording devices and patient conditions. 
Thus the color features are normalized using the first 30\% of each video, given the observation that the first 30\% is unlikely to contain heavy bleeding and should be of similar color and COF across cases. 
In detail, for each video, the mean color feature of the first 30\% is subtracted from the color feature of each frame. 
As for the semantic features, the ResNet-101~\cite{ResNet} pretrained on ImageNet is employed in each frame. 
Then the two types of features are concatenated. 
After feature extraction, the video is transformed into a feature sequence, denoted by $X \in \mathbb{R}^{T \times D}$. 
$T$ is the number of frames and $D$ is the feature dimension.

\noindent
{\bf Model.} 
A neural network model is designed for automated skill assessment, which consists of a score branch to evaluate frame quality and a weight branch to provide frame importance. 
Both of the branches are frame-wise multilayer perceptrons (MLP). 
Given the feature of a video frame, the score branch produces a score of this frame, while the weight branch outputs a frame weight. 
In this way, the feature sequence $X$ of the input video is transformed into a score sequence denoted by $A \in \mathbb{R}^{T \times 1}$ and a weight sequence represented by $U \in \mathbb{R}^{T \times 1}$. 
The weight branch is additionally followed by a softmax function so that the weights of all frames sum to one. 
Then the video-level score $q$ is obtained by the weighted sum of frame-level scores: $q = \sum_{t=1}^{T} U_t A_t$.

\noindent
{\bf Loss.} 
The loss function comprises a supervised regression loss and an unsupervised rank loss. 
The regression loss is a standard L1 loss: 
$L_{reg} = |y - q|$, 
where $y$ denotes the ground truth COF score and $q$ is the predicted score. 
In addition, we devise a rank loss based on the observation that the quality of COF decreases over time as the bleeding accumulates. It is assumed that the COF is better at the start of the surgery than in the end. As recommended by surgeons, we define the start section as the first 30\% of the surgery and the end section as 60\% to 90\%.
The last 10\% is not used for the end section because surgeons commonly clean the operating field thoroughly when finishing. Then the rank loss is proposed to enforce the score of the start section to be higher than the end section by a margin:
\begin{equation}
  L_{rank} = \mathrm{max}(0, 1 - (q_s - q_e))
\end{equation}
where $q_s = \sum_{t=1}^{0.3T} U_t A_t$ is the predicted score for the start section and $q_e = \sum_{t=0.6T}^{0.9T} U_t A_t$ is for the end section. 
This rank loss is only applied to the training cases with COF score no more than 3, since in good cases the COF quality might stay high through the whole surgery. 
Note that frame weights $U$ are normalized with softmax within the defined ranges when computing $q_s$ and $q_e$. 
Then the neural network is trained with the regression loss and the rank loss jointly. 




\begin{table}[t]
\begin{center}
\caption{Performance of COF Prediction} 
\label{table:table2}
\small
\begin{tabular}{M{3cm}|M{3cm}|M{2.5cm}|M{2.5cm}}
\hline
Method & Feature & PLCC & SROCC\\ 
\hline
Baseline & Red & 0.164 & 0.130 \\ 
Baseline & Saturation & 0.233 & 0.178 \\ 
Baseline & Duration & 0.229 & 0.191 \\ 
\hline
$\mathcal{L}_{reg}$ & Color + ResNet & 0.580 & 0.595 \\ 
$\mathcal{L}_{rank}$ & Color + ResNet & 0.433 & 0.457 \\ 
$\mathcal{L}_{reg}$ + $\mathcal{L}_{rank}$ & Color & 0.447 & 0.446 \\
$\mathcal{L}_{reg}$ + $\mathcal{L}_{rank}$ & ResNet & 0.622 & 0.601 \\
$\mathcal{L}_{reg}$ + $\mathcal{L}_{rank}$ & Color + ResNet & {\bf 0.641} & {\bf 0.647} \\
\hline
Junior Surgeon & - & 0.670 & 0.657 \\ 
Senior Surgeon & - & {\bf 0.880} & {\bf 0.869} \\ 
\hline
\end{tabular}
\end{center}
\end{table}

\begin{table}[t]
\begin{center}
\caption{Performance of Overall Skills Prediction}
\label{table:table3}
\small
\begin{tabular}{M{4.5cm}|M{3.3cm}|M{3.3cm}}
\hline
Method & PLCC (OTS/OPS) & SROCC (OTS/OPS)\\ 
\hline
No Proxy ($\mathcal{L}_{reg}$) & 0.46 / 0.18  & 0.47 / 0.18 \\    
No Proxy ($\mathcal{L}_{reg}$ + $\mathcal{L}_{rank}$) & 0.45 / 0.21  & 0.45 / 0.24  \\ 
With Proxy ($\mathcal{L}_{reg}$ + $\mathcal{L}_{rank}$) & {\bf 0.56} / {\bf 0.40}  & {\bf 0.55} / {\bf 0.41} \\ 
\hline
Junior Surgeon (COF) & 0.56 / 0.61 & 0.56 / 0.60 \\ 
Junior Surgeon (OTS/OPS) & 0.42 / 0.64 & 0.41 / 0.62 \\ 
Senior Surgeon (COF) & 0.74 / 0.74 & 0.73 / 0.73 \\ 
Senior Surgeon (OTS/OPS) & {\bf 0.82} / {\bf 0.84} & {\bf 0.82} / {\bf 0.83} \\ 
\hline
\end{tabular}
\end{center}
\end{table}

\section{Experiments}

The proposed method is evaluated on the newly introduced clinical dataset. 
We repeat three-fold cross-validation 15 times, with 45 runs in total. 
In each run, 38 videos are chosen randomly for training and the rest 19 videos are for testing. 
We report the Spearman's rank correlation coefficient (SROCC) and the Pearson linear correlation coefficient (PLCC) between the ground truth and the prediction, which are averaged over all 45 runs. 
Results of COF prediction and overall skills prediction are both presented.

\noindent
{\bf COF Prediction.} 
Results of COF prediction are provided in Table \ref{table:table2}. 
First, three baselines are tested, which simply depend on the mean red value, the mean saturation value, or duration of the video. 
Then ablation studies are performed to investigate the impact of feature and loss design. 
At last, human performances of senior and junior surgeons are computed by using annotations of each individual surgeon as the prediction and then averaging the results over juniors or seniors. 
Our full model surpasses the baselines and the ablated models, which is even comparable with junior surgeons.

\noindent
{\bf Overall Skills Prediction.} 
For overall skills, both OTS and OPS are predicted with or without the proxy of COF.
When with the proxy, the predicted overall skills are set the same as the predicted COF. 
When without proxy, same models are trained to regress overall skills directly by setting the OTS or OPS as the learning target $y$ in the loss. 
To facilitate the comparison between the proposed method and human performances, performances of surgeons are also computed with proxy (COF as overall skills) or without the proxy.
As the results in Table \ref{table:table3}, predicting overall skills via the proxy is better than directly regressing them without proxy, which further validates that COF is a good skill proxy. 
For OTS, the SROCC is 0.55, which is promising and comparable with the human performance of junior surgeons. 
For OPS, the SROCC is 0.41, which can be improved in the future by explicitly modeling the procedure steps. 
Note that previous works are not applicable due to different data modality and surgery setting. 

\begin{figure}[t]
\begin{center}
   \includegraphics[width=1.0\linewidth]{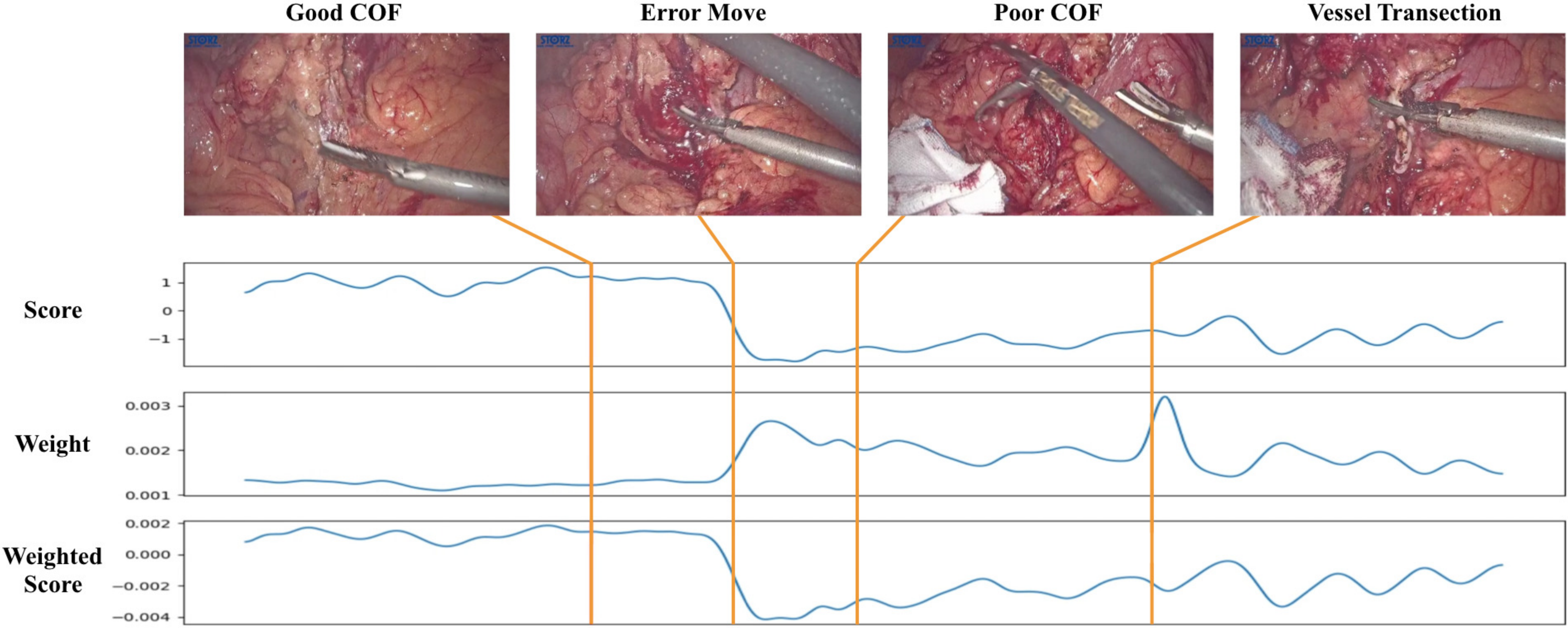}
\end{center}
   \caption{Frame-level COF scores for a test surgery. The scores have been normalized.} 
\label{fig:feedback}
\end{figure}

\noindent
{\bf Feedback.} 
As shown in Fig. \ref{fig:feedback}, the proposed method generates frame-level scores to indicate which parts are of low quality and frame-level weights to identify which parts should receive more attention. 
In this test example, the score drop (the 2nd image) corresponds to an error move that the surgeon needs to improve, and the increase of weight (the 4th image) corresponds to the vessel transection that the surgeon should perform with special care. 
Video demos are provided in the supplementary.
The quantification of such feedback is left for future research.

\section{Conclusion and Future Work} %
In this work, a clinical dataset for surgical skill assessment is compiled. 
Statistical analyses and empirical results both show that the clearness of operating field is a good skill proxy. 
A video-based objective model is proposed to predict overall skills via this proxy, which achieves promising results in experiments. 
The limitation of this work is that the dataset only includes gastrectomy procedures performed by a single surgeon. 
The generalizability to other procedures and across surgeons needs to be further validated.
Future works should also focus on 1) temporal structure modeling 2) incorporating more domain knowledge 3) explicitly modeling the surgery steps and human anatomy.

{\bf Acknowledgement.} 
This work was partially supported by National Basic Research Program of China (973 Program) under contract 2015CB351803, the Natural Science Foundation of China under contracts 61572042, 61527804 and 61625201. 
We also acknowledge the Clinical Medicine Plus X-Young Scholars Project and High-Performance Computing Platform of Peking University.








%
%
%
%

{\small
\bibliographystyle{splncs04}
\bibliography{mybib}

\begin{thebibliography}{10}
\providecommand{\url}[1]{\texttt{#1}}
\providecommand{\urlprefix}{URL }
\providecommand{\doi}[1]{https://doi.org/#1}

\bibitem{2013MICCAI}
Ahmidi, N., Gao, Y., B{\'e}jar, B., Vedula, S.S., Khudanpur, S., Vidal, R.,
  Hager, G.D.: String motif-based description of tool motion for detecting
  skill and gestures in robotic surgery. In: MICCAI (2013)

\bibitem{2015IJCARS}
Ahmidi, N., Poddar, P., Jones, J.D., Vedula, S.S., Ishii, L., Hager, G.D.,
  Ishii, M.: Automated objective surgical skill assessment in the operating
  room from unstructured tool motion in septoplasty. IJCARS  (2015)

\bibitem{JIGSAWS}
Ahmidi, N., Tao, L., Sefati, S., Gao, Y., Lea, C., Haro, B.B., Zappella, L.,
  Khudanpur, S., Vidal, R., Hager, G.D.: A dataset and benchmarks for
  segmentation and recognition of gestures in robotic surgery. IEEE TBE  (2017)

\bibitem{2019AnnS}
Azari, D.P., Frasier, L.L., Quamme, S.R.P., Greenberg, C.C., Pugh, C.M.,
  Greenberg, J.A., Radwin, R.G.: Modeling surgical technical skill using expert
  assessment for automated computer rating. Ann Surg  (2019)

\bibitem{NewEngland}
Birkmeyer, J.D., Finks, J.F., O'reilly, A., Oerline, M., Carlin, A.M., Nunn,
  A.R., Dimick, J., Banerjee, M., Birkmeyer, N.J.: Surgical skill and
  complication rates after bariatric surgery. N. Engl. J. Med  (2013)

\bibitem{2018CVPR}
Doughty, H., Damen, D., Mayol-Cuevas, W.: {Who's} better? {Who's} best?
  {Pairwise} deep ranking for skill determination. In: CVPR (2018)

\bibitem{2016MICCAI}
Ershad, M., Koesters, Z., Rege, R., Majewicz, A.: Meaningful assessment of
  surgical expertise: Semantic labeling with data and crowds. In: MICCAI (2016)

\bibitem{2018IJMRCAS}
Fard, M.J., Ameri, S., Darin~Ellis, R., Chinnam, R.B., Pandya, A.K., Klein,
  M.D.: Automated robot-assisted surgical skill evaluation: Predictive
  analytics approach. Int J Med Robotics Comput Assist Surg.  (2018)

\bibitem{ResNet}
He, K., Zhang, X., Ren, S., Sun, J.: Deep residual learning for image
  recognition. In: CVPR (2016)

\bibitem{GastrectomyBook}
Huang, C.M., Zheng, C.H.: Laparoscopic gastrectomy for gastric cancer: surgical
  technique and lymphadenectomy. Springer (2015)

\bibitem{2018MICCAI}
Ismail~Fawaz, H., Forestier, G., Weber, J., Idoumghar, L., Muller, P.A.:
  Evaluating surgical skills from kinematic data using convolutional neural
  networks. In: MICCAI (2018)

\bibitem{2018WACV}
Jin, A., Yeung, S., Jopling, J., Krause, J., Azagury, D., Milstein, A.,
  Fei-Fei, L.: Tool detection and operative skill assessment in surgical videos
  using region-based convolutional neural networks. In: WACV (2018)

\bibitem{OSATS}
Martin, J., Regehr, G., Reznick, R., Macrae, H., Murnaghan, J., Hutchison, C.,
  Brown, M.: Objective structured assessment of technical skill {(OSATS)} for
  surgical residents. British Journal of Surgery  (1997)

\bibitem{2010AnnS}
Richstone, L., Schwartz, M.J., Seideman, C., Cadeddu, J., Marshall, S.,
  Kavoussi, L.R.: Eye metrics as an objective assessment of surgical skill. Ann
  Surg  (2010)

\bibitem{2014ISBI}
Sharma, Y., Pl{\"o}tz, T., Hammerld, N., Mellor, S., McNaney, R., Olivier, P.,
  Deshmukh, S., McCaskie, A., Essa, I.: Automated surgical {OSATS} prediction
  from videos. In: ISBI (2014)

\bibitem{2017Survey}
Vedula, S.S., Ishii, M., Hager, G.D.: Objective assessment of surgical
  technical skill and competency in the operating room. Annu. Rev. Biomed. Eng
  (2017)

\bibitem{2018IJCARS_Wang}
Wang, Z., Fey, A.M.: Deep learning with convolutional neural network for
  objective skill evaluation in robot-assisted surgery. IJCARS  (2018)

\bibitem{2015TPAMI}
Zhang, Q., Li, B.: Relative hidden markov models for video-based evaluation of
  motion skills in surgical training. TPAMI  (2015)

\bibitem{2018IJCARS_Zia_2}
Zia, A., Essa, I.: Automated surgical skill assessment in {RMIS} training.
  IJCARS  (2018)

\bibitem{2018IJCARS_Zia_1}
Zia, A., Sharma, Y., Bettadapura, V., Sarin, E.L., Essa, I.: Video and
  accelerometer-based motion analysis for automated surgical skills assessment.
  IJCARS  (2018)

\bibitem{2016IJCARS}
Zia, A., Sharma, Y., Bettadapura, V., Sarin, E.L., Ploetz, T., Clements, M.A.,
  Essa, I.: Automated video-based assessment of surgical skills for training
  and evaluation in medical schools. IJCARS  (2016)

\end{thebibliography}
}

\end{document}